\pdfoutput=1

\documentclass[11pt]{article}

\usepackage{acl}

\usepackage{times}
\usepackage{latexsym}

\usepackage[disable]{todonotes}

\usepackage[T1]{fontenc}
\usepackage[utf8]{inputenc}

\usepackage{microtype}

\usepackage{makecell}

\usepackage{pifont}

\usepackage{color}

\newcommand{\forcamera}[1]{}
\usepackage{float}
\usepackage{makecell}

\newcommand{\caancora}[0]{\textsf{ca\_ancora}}%
\newcommand{\cspcedt}[0]{\textsf{cs\_pcedt}}%
\newcommand{\cspdt}[0]{\textsf{cs\_pdt}}%
\newcommand{\cuproiel}[0]{\textsf{cu\_proiel}}%
\newcommand{\deparcorfull}[0]{\textsf{de\_parcorfull}}%
\newcommand{\depotsdamcc}[0]{\textsf{de\_potsdam}}%
\newcommand{\engum}[0]{\textsf{en\_gum}}%
\newcommand{\enlitbank}[0]{\textsf{en\_litbank}}%
\newcommand{\enparcorfull}[0]{\textsf{en\_parcorfull}}%
\newcommand{\frdemocrat}[0]{\textsf{fr\_democrat}}%
\newcommand{\grcproiel}[0]{\textsf{grc\_proiel}}%
\newcommand{\hboptnk}[0]{\textsf{hbo\_ptnk}}%
\newcommand{\huszegedkoref}[0]{\textsf{hu\_szeged}}%
\newcommand{\hukorkor}[0]{\textsf{hu\_korkor}}%
\newcommand{\ltlcc}[0]{\textsf{lt\_lcc}}%
\newcommand{\nobokmaalnarc}[0]{\textsf{no\_bokmaalnarc}}%
\newcommand{\nonynorsknarc}[0]{\textsf{no\_nynorsk\-narc}}%
\newcommand{\plpcc}[0]{\textsf{pl\_pcc}}%
\newcommand{\rurucor}[0]{\textsf{ru\_rucor}}%
\newcommand{\esancora}[0]{\textsf{es\_ancora}}%
\newcommand{\tritcc}[0]{\textsf{tr\_itcc}}%

\newcommand{\baseline}[0]{\textsc{Baseline}}
\newcommand{\baselinegz}[0]{\textsc{Baseline-GZ}}
\newcommand{\baselinelast}[0]{\textsc{Baseline-2023}}
\newcommand{\bestlast}[0]{\textsc{Winner-2023}}

\def\MC#1#2{\multicolumn{#1}{c}{#2}}

\title{Findings of the Third Shared Task on Multilingual Coreference Resolution}

\author{
Michal Novák$^1$,
Barbora Dohnalová$^1$,
Miloslav Konopík$^2$,
Anna Nedoluzhko$^1$,
Martin Popel$^1$, \\
\textbf{Ondřej Pražák$^2$,
Jakub Sido$^2$,
Milan Straka$^1$,
Zdeněk Žabokrtský$^1$,
Daniel Zeman$^1$} \\[2mm]
$^1$ Charles University, Faculty of Mathematics and Physics, \\ Institute of Formal and Applied Linguistics, Prague, Czechia \\
\texttt{\{mnovak,nedoluzko,popel,straka,zabokrtsky,zeman\}@ufal.mff.cuni.cz}\\
\texttt{bdohnalova@matfyz.cz}\\[2mm]
$^2$
University of West Bohemia, Faculty of Applied Sciences, \\ Department of Computer Science and Engineering, Pilsen, Czechia \\
\texttt{\{konopik,ondfa,sidoj\}@kiv.zcu.cz}\\[2mm]
}

\usepackage{booktabs}
\newcommand{\ndatasets}[0]{21}
\newcommand{\nlanguages}[0]{15}
\newcommand{\nsystems}[0]{6} %

\usepackage{fancyhdr}
\fancypagestyle{officialbibref}{\fancyhf{}\fancyheadoffset[L,R]{50pt}\fancyhead[C]{This paper was published at \textbf{CRAC 2024} -- please cite the published version {\small\url{https://aclanthology.org/2024.crac-1.8}}.}\fancyfoot[C]{\normalsize\thepage}\addtolength{\topmargin}{-30pt}\addtolength{\headsep}{30pt}}
\begin{document}
\thispagestyle{officialbibref}\pagestyle{plain}
\maketitle

\begin{abstract}

The paper presents an overview of the third edition of the shared task on
multilingual coreference resolution, held as part of the CRAC 2024 workshop.
Similarly to the previous two editions, the participants were challenged to develop systems capable of identifying mentions and clustering them based on identity coreference.

This year's edition took another step towards real-world application by not providing participants with gold slots for zero anaphora, increasing the task's complexity and realism.
In addition, the shared task was expanded to include a more diverse set of languages, with a particular focus on historical languages.
The training and evaluation data were drawn from version 1.2 of the
multilingual collection of harmonized coreference resources CorefUD,
encompassing \ndatasets{} datasets across \nlanguages{} languages.
\nsystems{} systems competed in this shared task.

\end{abstract}

\section{Introduction}

The concept of a shared task dedicated to multilingual coreference resolution
began with SemEval-2010 \citep{recasens-etal-2010-semeval}, which included
seven languages, and CoNLL-2012 \cite{conll-2012}, which featured three
languages. In the Multilingual Coreference Resolution Shared Task at CRAC
2022 \cite{oursharedtask2022}, the scope was expanded to 10 languages, with
multiple datasets for some, using the CorefUD 1.0 collection
\cite{corefud2022lrec}. In the second edition of this shared task, held with
CRAC 2023, 12 languages were involved \cite{oursharedtask2023}. The
present paper details the third edition of this shared task, organized in
2024, once again in collaboration with CRAC.

This year's shared task introduces two significant changes compared to the previous edition.
First, there is an increased focus on zero mentions.
These zero mentions appear in 10 datasets for the following languages: Ancient
Greek, Catalan, Czech, Hungarian, Old Church Slavonic, Polish, Spanish, and
Turkish.
In the previous two editions of the shared task, zero mentions were technically present in the input (like any other mentions), which made the shared task's setting a bit artificial.
Now, requiring the participants not only to identify coreference relations but also to generate zeros in places relevant for coreference, makes the task closer to real-world scenarios (and harder).

Second, this year's shared task uses a newer version of CorefUD.
Compared to the previous version 1.1, CorefUD 1.2 comprises new languages and corpora.
Ancient Greek, Ancient Hebrew, and Old Church Slavonic have been added, further broadening the task's scope beyond Latin-script languages and toward those with significantly fewer resources.
Additionally, the introduction of LitBank for English extends the range of available domains by including novels with substantially longer documents.
These expansions aim to develop more robust solutions that are better suited for real-world applications.
Furthermore, updated versions of previously included resources, such as English-GUM and Turkish-ITCC, have been used.
The conversion of zeros in Polish-PCC has been considerably improved, and the conversion pipelines for
multiple other datasets have been refined too.

The rest of the paper is organized as follows. Section~\ref{sec:data}
discusses the changes in the shared task's data compared to the previous
edition. Section~\ref{sec:evaluation} outlines the evaluation metrics used in
the task, including both the primary and supplementary scores.
Section~\ref{sec:systems} details the baseline system and other participating
systems. Section~\ref{sec:results} presents a summary of the results and
Section~\ref{sec:conclusions} provides the conclusion.

\begin{table*}[!htb]
  \begin{center}
  \resizebox{\textwidth}{!}{
\begin{tabular}{@{}l rrrr rrrr rrrr @{}}\toprule
                              & \MC{4}{total number of}              & \MC{4}{entities}                 & \MC{4}{mentions}                  \\\cmidrule(lr){2-5}\cmidrule(lr){6-9}\cmidrule(lr){10-13}
document                      &        &         &         &         &  total & per 1k & \MC{2}{length} &   total & per 1k & \MC{2}{length} \\\cmidrule(lr){8-9}\cmidrule(lr){12-13}
                              &   docs &   sents &   words & empty n.&  count &  words &    max &  avg. &   count &  words &    max &  avg. \\\midrule
Ancient\_Greek-PROIEL         &     19 &   6,475 &  64,111 &   6,283 &   3,215 &     50 &    332 &   6.6 &  21,354 &    333 &     52 &   1.7 \\
Ancient\_Hebrew-PTNK          &     40 &   1,161 &  28,485 &       0 &     870 &     31 &    102 &   7.2 &   6,247 &    219 &     22 &   1.5 \\
Catalan-AnCora                &  1,298 &  13,613 & 429,313 &   6,377 &  17,558 &     41 &    101 &   3.6 &  62,417 &    145 &    141 &   4.8 \\
Czech-PCEDT                   &  2,312 &  49,208 & 1,155,755 &  35,654 &  49,225 &     43 &    236 &   3.4 & 168,055 &    145 &     79 &   3.6 \\
Czech-PDT                     &  3,165 &  49,428 & 834,720 &  21,808 &  46,628 &     56 &    172 &   3.3 & 154,905 &    186 &     99 &   3.1 \\
English-GUM                   &    217 &  12,147 & 211,920 &     115 &   8,270 &     39 &    131 &   4.4 &  36,733 &    173 &     95 &   2.6 \\
English-LitBank               &    100 &   8,560 & 210,530 &       0 &   2,164 &     10 &    261 &  10.8 &  23,340 &    111 &    129 &   1.6 \\
English-ParCorFull            &     19 &     543 &  10,798 &       0 &     188 &     17 &     38 &   4.4 &     835 &     77 &     37 &   2.1 \\
French-Democrat               &    126 &  13,057 & 284,883 &       0 &   7,162 &     25 &    895 &   6.5 &  46,487 &    163 &     71 &   1.7 \\
German-ParCorFull             &     19 &     543 &  10,602 &       0 &     243 &     23 &     43 &   3.7 &     896 &     85 &     30 &   2.0 \\
German-PotsdamCC              &    176 &   2,238 &  33,222 &       0 &     880 &     26 &     15 &   2.9 &   2,519 &     76 &     34 &   2.6 \\
Hungarian-KorKor              &     94 &   1,351 &  24,568 &   1,988 &   1,124 &     46 &     41 &   3.7 &   4,103 &    167 &     42 &   2.2 \\
Hungarian-SzegedKoref         &    400 &   8,820 & 123,968 &   4,857 &   4,769 &     38 &     36 &   3.2 &  15,165 &    122 &     36 &   1.6 \\
Lithuanian-LCC                &    100 &   1,714 &  37,014 &       0 &   1,087 &     29 &     23 &   4.0 &   4,337 &    117 &     19 &   1.5 \\
Norwegian-BokmaalNARC         &    346 &  15,742 & 245,515 &       0 &   5,658 &     23 &    298 &   4.7 &  26,611 &    108 &     51 &   1.9 \\
Norwegian-NynorskNARC         &    394 &  12,481 & 206,660 &       0 &   5,079 &     25 &     84 &   4.3 &  21,847 &    106 &     57 &   2.1 \\
Old\_Church\_Slavonic-PROIEL  &     26 &   6,832 &  61,759 &   6,289 &   3,396 &     55 &    134 &   6.5 &  22,116 &    358 &     52 &   1.5 \\
Polish-PCC                    &  1,828 &  35,874 & 538,885 &  18,615 &  22,143 &     41 &    135 &   3.7 &  82,706 &    153 &    108 &   1.9 \\
Russian-RuCor                 &    181 &   9,035 & 156,636 &       0 &   3,515 &     22 &    141 &   4.6 &  16,193 &    103 &     18 &   1.7 \\
Spanish-AnCora                &  1,356 &  14,159 & 458,418 &   8,112 &  19,445 &     42 &    110 &   3.6 &  70,663 &    154 &    101 &   4.8 \\
Turkish-ITCC                  &     24 &   4,732 &  55,358 &  11,584 &   4,019 &     73 &    369 &   5.4 &  21,569 &    390 &     31 &   1.1 \\
\bottomrule\end{tabular}
  }
  \caption{CorefUD~1.2 data sizes in terms of the total number of documents, sentences,
    words (i.e. non-empty nodes), empty nodes (empty words),
    coreference entities
    (total count, relative count per 1000 words, average and maximal length in number of mentions)
    and coreference mentions
    (total count, relative count per 1000 words, average and maximal length in number of words).
    All the counts are excluding singletons and for the concatenation of train+dev+test.
    Train/dev/test splits of these datasets roughly follow the 8/1/1 ratio.
    See Table~\ref{tab:sizes-train-dev} for details.
    }
  \label{tab:sizes}
  \end{center}
 \end{table*}

\section{Datasets}
\label{sec:data}

As in the previous years, the shared task takes its training and evaluation
data from the public part of the CorefUD collection
\cite{corefud2022lrec},\footnote{\url{https://ufal.mff.cuni.cz/corefud}} now
in its latest release
(1.2).\footnote{\url{http://hdl.handle.net/11234/1-5478}} The public edition
of CorefUD 1.2 consists of \ndatasets{} datasets for \nlanguages{} languages
(4 language families). Compared to CorefUD 1.1, which was used last year
\cite{oursharedtask2023}, there are 4 new datasets and 3 new languages
including one language (Ancient Hebrew) from a new language family. The new
datasets are Ancient Greek PROIEL, Old Church Slavonic PROIEL, Ancient Hebrew
PTNK, and English LitBank.
Beside adding these new datasets, most of the
``old'' datasets from CorefUD 1.1 were updated in various ways.
Table~\ref{tab:sizes} gives an overview of the datasets and their sizes.

\subsection{New Resources}

\textbf{Ancient Greek PROIEL} \citep[\grcproiel;][]{Haug2008CreatingAP} is a
collection of New Testament gospels from the PROIEL treebank. The main goal
of the PROIEL coreference annotation is to catch \textit{givenness}, i.e. how
readers determine the reference of nominal phrases. As a result,
referential noun phrases are annotated for identity coreference and bridging
relations, except relative pronouns and appositions. In addition to noun
phrases, zero anaphora for pro-dropped arguments is annotated, most often
unexpressed subjects. Due to the texts domain, special attention is paid
to the annotation of generic and other non-specific references. The original
annotation marks only mention heads, so the mention spans were determined
based on syntactic dependencies. Where possible, consecutive Bible chapters
were kept in the same document to preserve occasional cross-chapter
coreference links; however, coreference crossing training/dev/test boundaries
is lost. Manual morphosyntactic annotation from PROIEL was converted to the
UD scheme.

\textbf{Old Church Slavonic PROIEL} \citep[\cuproiel;][]{Haug2008CreatingAP}
includes Codex Marianus and selected chapters of Suprasliensis from the
PROIEL and TOROT treebanks. Coreference annotation follows the PROIEL
annotation guidelines, same as for Ancient Greek (see above). Manual
morphosyntactic annotation from PROIEL was converted to the UD scheme.

\textbf{Ancient Hebrew PTNK} \citep[\hboptnk;][]{swanson-etal-2024-towards}
contains portions of the Hebrew Bible as digitized and annotated in the
Biblia Hebraica Stuttgartensia. Entity and coreference annotation follows
guidelines similar to those of the English GUM corpus. Several high-frequency entities
have hundreds of mentions throughout the Bible (e.g., God, Abraham, Isaac or
Jacob); however, since the CorefUD 1.2 version of the resource uses chapters
as documents (which are then distributed between training/dev/test parts of
the data), coreference between chapters is not preserved. The current version
of the dataset also lacks annotation of zero mentions (their addition is
planned in the future, as Hebrew is a pro-drop language). Manual
morphosyntactic annotation was done natively in the UD scheme.

\textbf{English LitBank} \citep[\enlitbank;][]{Bamman2019AnAD} contains texts
from 100 literary novels of English-language fiction in LitBank. Compared to
other English corpora, the dataset contains longer texts with an average
length over 2000 words. Coreference annotation is close to the OntoNotes
coreference annotation style \cite{ontonotes-coref-manual} with several
significant changes such as explicit annotation of singletons and applying
coreference annotation to only the ACE categories (people, locations,
organizations, facilities, geopolitical entities, and vehicles, see
\citealp{walker2005ace}). Annotation of literary texts also demands for more
detailed insight into the identity phenomenon, thus near-identity or the
revelation of identity is paid more attention in the dataset. Morphosyntactic
annotation was predicted by UDPipe, as it was not part of the original
resource.
A coreference entity has on average 10.8 mentions,
 which is the highest number in CorefUD~1.2 (see Table~\ref{tab:sizes}).

\subsection{Updated Resources}

\paragraph{More data} The English GUM corpus (\engum) is now in its version
10, which has approximately 10\% more data. All the other datasets are the
same size as before (except for a few minor changes resulting from annotation
corrections).

\paragraph{Substantial changes} Re-implementation of conversion from
non-CorefUD formats and/or major revision of the annotation was applied to
French Democrat (\frdemocrat), Polish PCC (\plpcc), and Turkish ITCC
(\tritcc). Besides improved basic coreference annotation, in Polish and
Turkish this also involved a significant boost in annotation of zero mentions
(empty nodes), which are the theme of the present edition of the shared task.
Many changes were also applied to Czech (\cspdt, \cspcedt),
Catalan (\caancora) and Spanish (\esancora); here the changes affected both
the conversion of coreference and the manual morphosyntactic annotation in
UD.\footnote{More details on the changes can be found in the README files of
the individual corpora.}

\paragraph{New prediction of morphosyntax} Finally, for datasets that do not
come with manual morphosyntactic annotation, the UD relations, tags and
features were predicted with newer models for UDPipe (based on UD release
2.12). This involves all the remaining corpora except for the two Norwegian ones,
which did not change and have manual UD annotation.

\subsection{Zero mentions}
\label{sec:zeros}

\textit{Zero mention} refers to instances where a referent (typically the subject or object of a sentence) is implied but not explicitly mentioned in the text. Zero mention is common in pro-drop languages, where subject pronouns can be omitted because the verb conjugation often provides enough information to infer the subject.

In CorefUD, zero mentions are technically represented by \textit{empty nodes}, artificially inserted  into the UD trees in places where zero mentions are needed. Using this representation, a zero mention can be grouped with other mentions in a coreference chain to express coreference relations, fully analogously to overt (non-zero) mentions.

Languages differ substantially in what may be unexpressed. For example, Czech is considered a strongly pro-drop language and Russian is a partially pro-drop language, while English is not considered a pro-drop language. In addition, not only a subject pronoun but also an object or possessive pronoun can be dropped in some languages such as Hungarian. Another level of variability is caused by different design choices of authors of the original coreference resources; for example, some do annotate nominal ellipsis, while some do not. At this moment, harmonization of zero mentions is limited in CorefUD, and zero mentions from the original data resources are mostly preserved (i.e., captured by empty nodes).

In the previous two editions of this shared task, gold empty nodes (i.e., the slots for zero mentions) represented as empty nodes were available to participants both in the training and test data. That, however, was rather artificial, as zero mentions are by definition not overt in input texts. Hence their presence should be predicted too, as is the case in the current shared task.

\subsection{Data preprocessing and starting points}
\label{sec:data-preproc}

Compared to the public edition of CorefUD 1.2, the data provided for the shared task participants underwent slight adjustments.

\emph{Gold data} used for training and evaluation received a minor technical modification: the forms of empty nodes were removed.
This change was made to align the data more closely with the output of the baseline empty node prediction, which does not predict these forms (see Section~\ref{sec:baseline}).
Apart from this, the data remained consistent with the CorefUD~1.2 release, retaining manually annotated morpho-syntactic features (for datasets that originally included them), gold empty nodes, and gold coreference annotations.
While we made the gold train and dev sets available for download, the gold test set was kept secret and used exclusively within CodaLab for submissions evaluation.

\emph{Input data} were intended for processing by participants' systems and subsequent submission.
To better simulate a real-world scenario where no manual linguistic annotation is available, we removed the forms of empty nodes and replaced the original morpho-syntactic features with the outputs of UD~2.12 models across all datasets, including those with originally human-annotated features.
Additionally, the gold empty nodes and coreference annotations were removed.

Nevertheless, participants could choose from different \emph{starting points} for entering the shared task, with varying degrees of work required.
Depending on the chosen starting point, participants were provided with different levels of empty nodes' and coreference predictions from the baseline systems (see Section~\ref{sec:baseline}).
The three available starting points were:

\begin{enumerate}
    \item \emph{Coreference and zeros from scratch.}
    Participants were required to develop a system that resolves both coreference and predicts empty nodes potentially involved in zero anaphora.
    While this starting point is more challenging, it offers significant potential for gains.
    \item \emph{Coreference from scratch.}
    In this scenario, empty nodes were provided by the baseline system, allowing participants to focus solely on developing a coreference resolution system.
    Systems submitted in last year's edition could be applied to this starting point with some retraining.
    \item \emph{Refine the baseline.}
    Participants were given both empty nodes and coreference relations, as predicted by the baseline systems.
    This starting point is the simplest yet less flexible option.
\end{enumerate}

The input data preprocessing was performed on the dev and test sets.

\section{Evaluation Metrics}
\label{sec:evaluation}

The systems participating in the shared task are evaluated with the CorefUD scorer.
Similarly to the last year's edition, the primary evaluation score is the CoNLL F1 score with head mention matching and singletons excluded.
As gold and predicted zero mentions are no longer guaranteed to match one-to-one, we introduce the dependency-based method to align them.
Furthermore, we calculate several other supplementary scores to compare the shared task submissions.

\paragraph{Official scorer}

We use the CorefUD scorer%
\footnote{\url{https://github.com/ufal/corefud-scorer}}
in its version from May 2024 to evaluate the submissions of the participants.
It has been upgraded to build on the Universal Anaphora (UA) scorer 2.0 \citep{ua-scorer-2.0} instead of the UA scorer 1.0 \citep{UA-scorer-2022}.
Besides the features that had been an integral part of the older CorefUD scorer and were newly introduced to the UA scorer 2.0, e.g., Mention Overlap Ratio \citep[MOR;][]{oursharedtask2022}, anaphor-level evaluation of zeros, support for discontinuous mentions and the CorefUD 1.0 file format, the upgrade fixed a bug in partial matching method and introduced the linear method of matching zero mentions.
Naturally, it still allows to take advantage of the implementations of all generally used coreferential measures with no modifications.
Unlike the UA scorer, the CorefUD scorer provides support for head match and newly for dependency-based method of matching zero mentions.

\paragraph{Mention matching}
Due to shortcomings of using \emph{exact} and \emph{partial} mention matching (see \citet{oursharedtask2023} for details), we arrived at the decision to use the \emph{head match} method in the primary metrics last year.
Gold and predicted mentions are considered matching if their heads%
\footnote{Note that gold mention heads in the CorefUD data were determined from the dependency tree using the Udapi block \texttt{corefud.MoveHead}.}
correspond to identical tokens.
Full spans are ignored, except for the case of multiple mentions with the same head in order to disambiguate between them.

\paragraph{Matching of zeros}
However, none of the matching methods can be any longer applied to empty nodes.
As in this year the participants are expected to predict empty nodes involved in zero anaphora, they are not guaranteed to align one-to-one with the gold empty nodes.
They can be missing, spurious, or predicted at different surface positions within the sentence, yet playing the same role.

We thus introduce the \emph{dependency-based method of matching zero mentions}.
It looks for the matching of zeros within the same sentence that maximizes the F-score of predicting dependencies of zeros in the enhanced dependency graph.%
\footnote{Stored in the \textsc{DEPS} field of the CoNLL-U format.}
Specifically, the task is cast as searching for a one-to-one matching in a weighted bipartite graph (with gold and predicted mentions as the two partitions) to maximize the total sum of weights in the matching.
Each candidate pair (gold zero mention -- predicted zero mention) is weighed with a non-zero score only if the two mentions belong to the same sentence.
The score is then calculated as a weighted sum of two features:

\begin{itemize}
    \item the F-score of the gold zero dependencies recognized in the predicted zero, considering both parent and dependency type assignments (weighted by a factor of 10);
    \item the F-score of the gold zero dependencies recognized in the predicted zero, considering only parent assignments (weighed by a factor of 1).
\end{itemize}

The scoring mechanism prioritizes the exact assignment of both parents and types.
Nevertheless, it is ensured to sufficiently work even if the predictions contain no dependency type assignments.

This matching strategy differs to the linear matching of zeros presented by \citet{ua-scorer-2.0}, which aligns the zeros only if their word indices%
\footnote{Stored in the \textsc{ID} field of the CoNLL-U format.}
are identical.
Such matching may thus fail if the zero is predicted at different surface position or if only one of the multiple zeros with the same parent is predicted.

\paragraph{Primary score}
Following the best practices for coreference resolution, we utilize the CoNLL $F_1$ score \citep{CoNLL-MELA-score,pradhan-etal-2014-scoring} as the primary evaluation score.
It is an unweighted average of the $F_1$ scores of three coreference metrics: MUC \citep{MUC-score}, B$^3$ \citep{Bcubed-score} and CEAF-e \citep{CEAF-score}.
The final ranking of participating submissions is then based on a macro-average of CoNLL $F_1$ scores over all datasets in the CorefUD test collection.

\paragraph{Supplementary scores}
Besides the primary CoNLL $F_1$ score, we report alternative versions of this score using different ways of mention matching:
partial match%
\footnote{The partial-match setup was used in the primary metrics in the first edition of the shared task \citep{oursharedtask2022}.}
and exact match.
Furthermore, we calculate the
primary metrics using the head-match for all mentions including singletons.

We also report the systems' performance in terms of the coreference metrics that contribute to the CoNLL score as well as other standard measures, e.g. BLANC \citep{BLANC-score} and LEA
\citep{LEA-score}.
We employ the MOR score to evaluate the quality of mention matching, while ignoring the assignment of mentions to coreferential entities.
Moreover, this year, it is particularly interesting to analyze the performance of the systems on zero anaphora.
To this end, we use the anaphor-decomposable score for zeros \citep{oursharedtask2022}, which is an application of the scoring schema proposed by \citet{tuggener-2014-coreference}.

\section{Participating Systems}
\label{sec:systems}

\subsection{Baseline}
\label{sec:baseline}

This year, two baseline systems are provided: one for predicting empty nodes as slots for zero anaphora, and another for coreference resolution.

\paragraph{Empty Nodes Prediction Baseline} Predicting empty nodes is a novel task in this year's shared task. To accommodate participants who want to focus solely on coreference resolution, we provide a baseline for predicting empty nodes. We release the source code,\footnote{\label{ftn:empty-nodes-baseline}\url{https://github.com/ufal/crac2024_zero_nodes_baseline}} the trained multilingual model,\footnote{\url{https://www.kaggle.com/models/ufal-mff/crac2024_zero_nodes_baseline/}} and development and testing data with predicted empty nodes.

The baseline model architecture is as follows. Every sentence is processed independently, and its words are split into subwords by the XLM-RoBERTa tokenizer~\citep{conneau-etal-2020-unsupervised}. The subwords are passed through the XLM-RoBERTa large pretrained model, and the embeddings of the first subword of every word are utilized as the word representations. Then, two candidate representations for every word are generated, by (1)~passing the word representations through a ReLU-activated 2k-unit dense layer, a dropout layer and a 768-unit dense layer; (2)~concatenating the described outputs with the original word representations and passed through an analogous dense-dropout-dense module. Each candidate representation might generate an empty node, whose dependency head would be the word generating the candidate. The candidate representations are processed by three heads, each first applying a 2k-unit dense layer, ReLU, and dropout: (1)~a binary classification head predicting whether the candidate is an empty node or not, (2)~word-order prediction head implemented using self-attention selecting the word after which the empty node should be added, and (3)~dependency relation prediction head, which first concatenates the candidate representation and the representation of the word most probable according to the word-order prediction head, and then predicts the dependency relation.

The model was trained on a combination of all languages containing empty nodes, sampling every language proportionally to the square root of its size. Further details and used hyperparameters are available in the source code repository.$^{\ref{ftn:empty-nodes-baseline}}$

The performance of the empty nodes prediction baseline is quantified in Table~\ref{tab:zeros_baseline} using precision, recall, and F1 score, where a predicted empty node is considered correct if its dependency head, dependency relation, and word order are all correct.

\begin{table}[t]
  \centering
  \begin{tabular}{lccc} \toprule
  \textbf{Language} & \textbf{Recall} & \textbf{Precision} & \textbf{F1} \\ \midrule
\caancora      & 91.01 & 92.32 & 91.66 \\
\cspcedt       & 59.84 & 78.22 & 67.81 \\
\cspdt         & 71.56 & 81.47 & 76.19 \\
\cuproiel      & 78.76 & 81.61 & 80.16 \\
\esancora      & 91.92 & 92.04 & 91.98 \\
\grcproiel     & 86.58 & 90.29 & 88.39 \\
\hukorkor      & 60.21 & 74.68 & 66.67 \\
\huszegedkoref & 89.52 & 91.93 & 90.71 \\
\plpcc         & 91.61 & 87.50 & 89.51 \\
\tritcc        & 93.81 & 79.05 & 85.80 \\
  \bottomrule
  \end{tabular}
  \caption{Empty nodes prediction baseline performance on the development sets of CorefUD 1.2 languages containing empty nodes. An empty node is considered correct if it has the correct dependency head, dependency relation, and word order.}
  \label{tab:zeros_baseline}
\end{table}

\paragraph{Coreference Resolution Baseline}

The baseline for coreference resolution is the same as in the two previous years. It is a multilingual end-to-end neural coreference resolution by \cite{prazak-etal-2021-multilingual}. The model is the adaptation of the standard end-to-end neural coreference resolution system originally proposed by \citet{lee-etal-2017-end}. The model iterates over all possible spans up to the maximum length and predicts the antecedent for each potential span directly. Because it does not predict the mentions in the separate step, it should be sufficient for the datasets where singletons are not annotated. The baseline coreference model uses mBERT base as an encoder.

\subsection{System Submissions}\label{sec:system-submissions}
This year,
six systems were submitted to the shared task by the following four teams:
 DFKI\_TR,\footnote{DFKI = Deutsches Forschungszentrum für Künstliche Intelligenz (German Research Center for Artificial Intelligence).
 The DFKI-CorefGen system was submitted to CodaLab by user ``natalia\_s''.
 }
 ÚFAL CorPipe,\footnote{ÚFAL = Ústav formální a aplikované lingvistiky (Institute of Formal and Applied Linguistics).
 The ÚFAL CorPipe team submitted 3 systems:
  CorPipe, CorPipe-2stage and CorPipe-single,
  by CodaLab users ``straka'', ``straka-twostage'' and ``straka-single-multilingual-model'', respectively.
 }
 UWB\footnote{UWB = University of West Bohemia. The Ondfa system was submitted to CodaLab by user ``ondfa''.}
 and Ritwikmishra.\footnote{The Ritwikmishra system was submitted to CodaLab by user ``ritwikmishra''.}
Some of the files produced by the Ritwikmishra system were not valid CoNLL-U
 and the scorer failed, thus resulting in zero F1 for these datasets (see Table~\ref{tab:all-langs}).
We applied an automatic correction\footnote{Mostly moving Entity annotations from multi-word tokens
 (where they are forbidden) to the words.}
 and call the resulting system RitwikmishraFix.
The tables with results in Section~\ref{sec:results} also include the baseline system (\baseline)
as described in Section~\ref{sec:baseline}
and the same baseline system applied on gold empty nodes (\baselinegz).
The total number of systems compared is thus 9.

The following descriptions are based on the information provided by the respective participants in an online questionnaire.
Basic properties of the systems are also summarized in Table~\ref{tab:comparison}.

\paragraph{DFKI-CorefGen}
The DFKI-CorefGen system performs mention identification and co-reference resolution
jointly, treating both tasks as text generation. Given a piece of text, the system
identifies all mentions and groups them into clusters by marking the mentions with
square brackets accompanied by cluster identifiers. The approach resolves co-reference
incrementally, processing each new sentence to find mentions and cluster them, while
also correcting cluster assignments in the previous context if needed.

To train the model, DFKI-CorefGen applies prefix tuning using OpenPrompt \citep{ding2021openprompt}.
The system utilizes multilingual T5 base \citep{xue-etal-2021-mt5} as the foundation model.
During training, the pre-trained model is kept frozen, and only the prefix component is tuned.

\paragraph{CorPipe-2stage}
CorPipe-2stage is a minor evolution of the system implemented in the previous year~\citep{sharedtask-straka-2022}. It combines the baseline provided by the shared task organizers for the prediction of zeros, followed by the last year's version of CorPipe, which first predicts the mentions and then the links among them using a single pre-trained Transformer encoder. Three model variants are trained, based on either mT5-large, InfoXLM-large, or mT5-xl. For every variant, 7 multilingual models are trained on a combination of all the treebanks, differing only in random initialization. The treebanks are sampled proportionally to the square root of their size, and most hyperparameters are taken from the last year's CorPipe. Then, for each treebank, the best-performing checkpoints are selected from the shared pool of checkpoints and ensembled.

\paragraph{CorPipe}
Contrary to the CorPipe-2stage submission using two Transformer encoders, the submission CorPipe predicts the zero mentions directly from the words, jointly with the nonzero mention prediction and the link prediction. It uses the same approach of 3 Transformer
encoder variants, 7 multilingual models per variant, and ensemble selection for
each treebank.

\paragraph{CorPipe-single}
CorPipe-single uses the same checkpoint pool as the CorPipe system, but it chooses a single mT5-large-based model for prediction on all treebanks.

\paragraph{Ondfa}
The Ondfa system extends the baseline system and participant systems from previous years \cite{prazak-konopik-2022-end}.
The approach involves initially training a joint cross-lingual model (XLM-R-large, mT5-xxl)
for all datasets. Subsequently, the model is fine-tuned for each dataset separately,
using LORA in the case of mT5.

Mentions are newly represented only with their headwords
(except for \cspcedt{} and \ltlcc{}, where multiword mentions were allowed),
which has been shown to improve
the primary metric (head-match) results
on the dev sets.
Syntax trees are also incorporated
as features into the model. The UWB team also modified their model to handle singletons.

\paragraph{Ritwikmishra}
This submission reuses the TransMuCoRes system from \citep{mishra-etal-2024-multilingual}, which is a fine-tuned \emph{wl-coref} architecture \citep{dobrovolskii-2021-word} built on top of the XLM-R-base model.
This system is applied in a zero-shot manner
on both the development and test sets.

\begin{table*}[!t]
\centering
\begin{tabular}{l l l l} \hline
\textbf{Name} & \textbf{Baseline} & \textbf{Starting point} & \textbf{Official data} \\ \hline
DFKI-TR & No & From scratch & Yes \\
CorPipe & No & From scratch & Yes \\
CorPipe-single & No & From scratch & Yes \\
CorPipe-2stage & Prediction of zeros & Coreference from scratch & Yes \\
Ondfa & Coref. resolution & Coreference from scratch & Yes \\
Ritwikmishra & No & Coreference from scratch & No (TransMuCoRes) \\
\hline \end{tabular}

\vspace{1em}
\begin{tabular}{l l l l} \hline
\textbf{Name} & \textbf{Pretrained model} & \textbf{Model size} & \textbf{Seq. length} \\ \hline
DFKI-TR & mT5-base & 580M + 3.4M & 512 subwords \\
CorPipe & \makecell[tl]{mT5-large,\\mT5-xl,\\InfoXLM-large} & \makecell[tl]{3.7B+280M (3-model\\~~ensemble,\\~~average)} & \makecell[tl]{2560 for mT5,\\512 for InfoXLM,\\512 during training} \\
CorPipe-single & mT5-large & 538M+57M & \makecell[tl]{2560 during prediction,\\512 during training} \\
CorPipe-2stage & \makecell[tl]{mT5-large,\\mT5-xl,\\InfoXLM-large} & \makecell[tl]{5.1B+400M (5-model\\~~ensemble,\\~~average)} & \makecell[tl]{2560 for mT5,\\512 for InfoXLM,\\512 during training} \\
Ondfa & \makecell[tl]{XLM-R-large,\\mT5-xxl} & \makecell[tl]{550M + 20M (xlmr),\\5.7B + 70-400M (mt5)} & \makecell[tl]{512, 2048, 4096 \\2048, 4096} \\
Ritwikmishra & XLM-R-base & 270M + 4.3M & variable \\
\hline \end{tabular}

\vspace{1em}
\begin{tabular}{l l l l} \hline \textbf{Name} & \textbf{Tuned per lang.?} & \textbf{Batch size} & \textbf{Tuned hyperparameters} \\ \hline
DFKI-TR & No & 1 & Not specified \\
CorPipe & Yes (21 models) & 8, 12 & Model variant (rest taken from 2023) \\
CorPipe-single & No & 8 & Taken from 2023 \\
CorPipe-2stage & Yes (21 models) & 8, 12 & Model variant (rest taken from 2023) \\
Ondfa & Yes & 1 doc & LORA rank (rest taken from 2023) \\
Ritwikmishra & No & 8 & None \\ \hline
\end{tabular}
\caption{The table compares properties of systems participating in the task.
The systems are ordered alphabetically.
The shortcuts in the headings are defined as follows: \textbf{Name} is the name of the submission, \textbf{Baseline}: what type of baseline the system builds on (see Section \ref{sec:baseline}).
\textbf{Starting point}: the chosen starting level out of the three possible ones as listed in Section~\ref{sec:data-preproc}, \emph{From scratch} denotes the \emph{Coreference and zeros from scratch} starting point.
\textbf{Official data}: Use of CorefUD 1.2 public edition for training, \textbf{Tuned per lang.?} indicates whether participants tuned their model for each language or not. \textbf{Model size}: The model size is split between the Pretrained model size and the size of the added head. \textbf{variable} means various settings depending on features and architecture.}

\label{tab:comparison}
\end{table*}

\subsection{System Comparison}

Most of the systems, including DFKI-CorefGen and the CorPipe variants, developed their approaches completely from scratch.
However, CorPipe-2stage, Ritwikmishra, and Ondfa utilized the provided baseline predictions of
empty nodes (the \emph{Coreference from scratch} starting point).
Additionally, Ondfa built upon the baseline coreference resolution system, but no submission was based solely on the baseline predictions (the \emph{Refine the baseline} starting point).

The systems leveraged various pre-trained models: DFKI-CorefGen employed mT5-base \citep{xue-etal-2021-mt5}; the CorPipe variants used combinations of encoder blocks from mT5-large, mT5-xl \citep{xue-etal-2021-mt5}, and InfoXLM-large \citep{chi-etal-2021-infoxlm}; Ondfa utilized XLM-R-large \citep{conneau-etal-2020-unsupervised} and mT5-xxl \citep{xue-etal-2021-mt5}; and Ritwikmishra opted for XLM-R-base \citep{conneau-etal-2020-unsupervised}.

Model sizes varied significantly, ranging from around 600M parameters for DFKI-CorefGen and Ritwikmishra to 6.1B for Ondfa's mT5-xxl model. The CorPipe systems distinguished themselves by employing ensemble methods with multiple models. Language-specific tuning was another point of differentiation: CorPipe, CorPipe-2stage, and Ondfa fine-tuned their models for individual languages, while DFKI-CorefGen, CorPipe-single, and Ritwikmishra maintained a single multilingual model approach.

Regarding training data, most systems utilized the official CorefUD 1.2 public edition. Ritwikmishra, however, diverged from this trend by using the TransMuCoRes dataset \citep{mishra-etal-2024-multilingual}.

\section{Results and Comparison}
\label{sec:results}

\subsection{Main Results}
The main results are summarized in Table~\ref{tab:main-results}.
The CorPipe-2stage system is the best one according to the official primary metric
 (head-match excluding singletons) as well as according to three alternative metrics:
 partial-match excluding singletons (which was the primary metric in 2022),
 exact-match excluding singletons
 and head-match including singletons.
All four metrics result in the same ordering of systems
 with a single exception of the Ondfa system,
 which is the sixth best according to exact-match,
 but the fourth best according to other metrics.
This is caused by the fact that for all but two datasets
 (cf. description of Ondfa in Section~\ref{sec:system-submissions}),
 Ondfa predicted only the head word and the span was always just this single word.

The third edition of the shared task is also a good time to look into how the state of the art in multilingual coreference resolution develops.
However, the results are not directly comparable across the years as the CorefUD collection has grown and some details of the shared task have changed over the years.
The baseline system has not fundamentally changed, set aside that it has been trained on slightly different data.
We can thus compare the relative improvement of the best system over the baseline.
As shown in Table~\ref{tab:main-results}, while the gain over the baseline was 31\% last year, this year it is 39\%.

\begin{table*}\centering
\begin{tabular}{@{}l r r@{~~}l r@{~~}l r@{~~}l @{}}\toprule
              & \MC{5}{excluding singletons} & \MC{2}{with singletons}\\\cmidrule(lr){2-6}\cmidrule(l){7-8}
\bf system    & \bf head-match & \MC{2}{\bf partial-match} & \MC{2}{\bf exact-match} & \MC{2}{\bf head-match}\\\midrule
CorPipe-2stage & \bf 73.90 & \bf 72.19 & (-1.71) & \bf 69.86 & (-4.04) & \bf 75.65 & (+1.75)\\
CorPipe       &     72.75 &     70.30 & (-2.45) &     68.36 & (-4.39) &     74.65 & (+1.90)\\
CorPipe-single &     70.18 &     68.02 & (-2.16) &     66.07 & (-4.11) &     71.96 & (+1.78)\\
Ondfa         &     69.97 &     69.82 & (-0.15) &     40.25 & (-29.72) &     70.67 & (+0.69)\\
\baselinegz   &     54.60 &     53.95 & (-0.65) &     52.63 & (-1.97) &     47.89 & (-6.71)\\
\baseline     &     53.16 &     52.48 & (-0.68) &     51.26 & (-1.90) &     46.45 & (-6.71)\\
DFKI-CorefGen &     33.38 &     32.36 & (-1.02) &     30.71 & (-2.68) &     38.65 & (+5.26)\\
RitwikmishraFix &     30.63 &     32.21 & (+1.58) &     28.27 & (-2.35) &     27.05 & (-3.58)\\
Ritwikmishra  &     16.47 &     16.65 & (+0.17) &     14.16 & (-2.31) &     15.42 & (-1.06)\\
\midrule
\bestlast{}     &    74.90 &      73.33 & (-1.57) &     71.46 & (-3.44) &     76.82 & (+1.91)\\
\baselinelast{} &    56.96  &     56.28 & (-0.68) &     54.75 & (-2.21) &     49.32 & (-7.64)\\
\bottomrule\end{tabular}

\caption{Main results: the CoNLL metric macro-averaged over all datasets.
The table shows the primary metric (head-match excluding singletons)
 and three alternative metrics:
 partial-match excluding singletons,
 exact-match excluding singletons and
 head-match with singletons.
A difference relative to the primary metric is reported in parenthesis.
The best score in each column is in bold.
The systems are ordered by the primary metric.
The last two rows showing the winner and baseline results from CRAC~2023
 are copied from the last year Findings \citep{oursharedtask2023},
 and thus are not directly comparable with the rest of the table
 because both the test and training data have been changed (CorefUD~1.1 vs. 1.2).
Similar notes apply to the following tables.
}
\label{tab:main-results}
\end{table*}

Table~\ref{tab:secondary-metrics} shows recall, precision, and F1 for six
metrics.
The F1 scores of the first five metrics (MUC. B$^3$, BLANC, and LEA)
 result in the same ordering of systems (same as the primary metric)
 except for RitwikmishraFix, which is slightly better than DFKI-CorefGen in BLANC and LEA.
Most of the systems have higher precision than recall for all the metrics,
 but the highest disbalance is in the \baseline{} system.
CorPipe* are the only systems that have higher recall than precision
 at least for CEAF-e (but other metrics have similar precision and recall).

The MOR metric (mention overlap ratio) measures only the mention matching quality,
 while ignoring the coreference, but even then
 the ordering of systems is similar to the primary metric
 (Ondfa is the fourth worst according to MOR,
  again because it does not predict full spans for most datasets).

\begin{table*}\centering
\begin{tabular}{@{}l cccccc @{}}\toprule
\bf system          & \bf MUC &\bf B$^3$ &\bf CEAF-e &\bf BLANC &\bf LEA &\bf MOR\\\midrule
CorPipe-2stage &  {\bf 79} / {\bf 81} / {\bf 80}  &  {\bf 69} / {\bf 74} / {\bf 71}  &  {\bf 71} / {\bf 70} / {\bf 70}  &  {\bf 67} / {\bf 73} / {\bf 70}  &  {\bf 66} / {\bf 71} / {\bf 68}  &       78  /      82  / {\bf 80} \\
CorPipe       &       79  /      80  /      79   &       69  /      72  /      70   &       71  /      68  /      69   &       67  /      72  /      69   &       65  /      69  /      67   &       78  /      80  /      79  \\
CorPipe-single &       77  /      76  /      77   &       68  /      67  /      67   &       69  /      66  /      67   &       66  /      66  /      66   &       64  /      63  /      64   &  {\bf 79} /      77  /      77  \\
Ondfa         &       75  /      81  /      78   &       64  /      72  /      67   &       64  /      67  /      65   &       62  /      71  /      65   &       61  /      69  /      64   &       41  / {\bf 87} /      54  \\
\baselinegz &       56  /      75  /      63   &       43  /      63  /      50   &       46  /      57  /      50   &       41  /      63  /      48   &       39  /      58  /      46   &       49  /      86  /      61  \\
\baseline     &       54  /      73  /      62   &       41  /      62  /      49   &       44  /      56  /      49   &       39  /      62  /      46   &       37  /      57  /      44   &       48  /      85  /      60  \\
DFKI-CorefGen &       37  /      52  /      41   &       26  /      38  /      29   &       25  /      42  /      30   &       21  /      39  /      23   &       21  /      31  /      23   &       43  /      71  /      50  \\
RitwikmishraFix &       33  /      50  /      36   &       26  /      43  /      28   &       27  /      37  /      29   &       24  /      39  /      24   &       24  /      39  /      25   &       30  /      65  /      36  \\
Ritwikmishra  &       18  /      31  /      18   &       15  /      27  /      15   &       15  /      22  /      16   &       13  /      23  /      12   &       13  /      25  /      13   &       17  /      38  /      20  \\
\bottomrule\end{tabular}

\caption{Recall / Precision / F1 for individual secondary metrics.
All scores macro-averaged over all datasets.
}
\label{tab:secondary-metrics}
\end{table*}

Table~\ref{tab:all-langs} shows the primary metric (CoNLL F1 head-match) for individual datasets.
The winner (CorPipe-2stage) is the best system for 15 out of \ndatasets{} datasets,
so the results are more diverse than last year,
when the winner (CorPipe) was the best system across all datasets and languages.
Interestingly, there is a substantial improvement of all systems on \tritcc{}
 relative to the last year
 (\baselinegz{} 51.16\% this year vs. \baselinelast{}=22.75\% last year;
 the winner has 68.18 this year vs. 55.63 last year).
This is due to the fixes in the dataset and possibly because zero anaphora was newly introduced in the source corpus \citep{pamayTurkishCR24}.

\newcommand*\rot{\rotatebox{90}}
\begin{table*}\centering
\scriptsize
\resizebox{\textwidth}{!}{
\begin{tabular}{@{}lr@{~~~} r@{~~~}r@{~~~}r@{~~~}r@{~~~}r@{~~~}r@{~~~}r@{~~~}r@{~~~}r@{~~~}r@{~~~}r@{~~~}r@{~~~}r@{~~~}r@{~~~}r@{~~~}r@{~~~}r@{~~~}r@{~~~}r@{~~~}r@{~~~}r@{}}\toprule
\bf system          & \rot\caancora & \rot\cspcedt & \rot\cspdt & \rot\cuproiel & \rot\deparcorfull & \rot\depotsdamcc & \rot\engum & \rot\enlitbank & \rot\enparcorfull & \rot\esancora & \rot\frdemocrat & \rot\grcproiel & \rot\hboptnk & \rot\hukorkor & \rot\huszegedkoref & \rot\ltlcc & \rot\nobokmaalnarc & \rot\nonynorsknarc & \rot\plpcc & \rot\rurucor & \rot\tritcc\\\midrule
CorPipe-2stage &    82.22 &\bf 74.85 &\bf 77.18 &\bf 61.58 &    69.53 &    71.79 &\bf 75.66 &\bf 79.60 &    68.89 &\bf 82.46 &    68.16 &\bf 71.34 &\bf 72.02 &    63.17 &\bf 69.97 &\bf 75.79 &\bf 79.81 &\bf 78.01 &\bf 78.50 &\bf 83.22 &\bf 68.18\\
CorPipe       &    81.02 &    73.71 &    75.84 &    60.72 &\bf 71.68 &    71.45 &    74.61 &    79.10 &\bf 69.75 &    80.98 &\bf 68.77 &    68.53 &    70.86 &    60.32 &    68.12 &    75.78 &    79.55 &    77.52 &    77.03 &    83.09 &    59.37\\
CorPipe-single &    80.42 &    72.82 &    74.82 &    57.11 &    61.62 &    67.02 &    74.39 &    78.08 &    58.61 &    79.75 &    67.89 &    66.01 &    67.18 &    60.09 &    67.32 &    75.19 &    78.92 &    76.60 &    75.20 &    81.21 &    53.43\\
Ondfa         &\bf 82.46 &    70.82 &    75.80 &    54.97 &    71.40 &\bf 71.91 &    70.53 &    74.15 &    55.58 &    81.94 &    62.69 &    61.64 &    61.56 &\bf 64.86 &    69.26 &    71.97 &    74.51 &    72.07 &    76.34 &    80.47 &    64.49\\
\baselinegz &    69.59 &    68.93 &    66.15 &    27.56 &    47.21 &    55.65 &    63.18 &    63.54 &    33.08 &    70.64 &    53.62 &    31.87 &    24.60 &    41.65 &    54.64 &    62.00 &    64.96 &    63.70 &    67.00 &    65.83 &    51.16\\
\baseline     &    68.32 &    64.06 &    63.83 &    24.51 &    47.21 &    55.65 &    63.19 &    63.54 &    33.08 &    69.58 &    53.62 &    28.76 &    24.60 &    35.14 &    54.51 &    62.00 &    64.96 &    63.70 &    66.24 &    65.83 &    44.05\\
DFKI-CorefGen &    34.77 &    32.89 &    30.88 &    22.52 &    23.07 &    45.85 &    35.49 &    46.59 &    32.69 &    37.76 &    36.34 &    25.87 &    37.96 &    23.53 &    33.85 &    42.73 &    37.92 &    35.69 &    27.19 &    47.79 &     9.65\\
RitwikmishraFix &    27.05 &     0.00 &     0.00 &     6.79 &    25.35 &    48.90 &    48.64 &    61.47 &    53.12 &    30.04 &    43.63 &     5.60 &     0.12 &    33.40 &    30.28 &    44.31 &    56.41 &    53.17 &     0.00 &    53.89 &    20.97\\
Ritwikmishra  &     0.00 &     0.00 &     0.00 &     6.79 &    25.35 &    48.90 &     0.00 &     0.00 &    53.12 &     0.00 &    43.72 &     5.60 &     0.09 &    33.40 &    30.32 &    44.78 &     0.00 &     0.00 &     0.00 &    53.88 &     0.00\\
\midrule
\baselinelast{}           &    65.26 &    67.72 &    65.22 &    -- & 44.11 &    57.13 &    63.08 &    -- &   35.19 &    66.93 &    55.31 &   -- &   -- &    40.71 &    55.32 &    63.57 &    65.10 &    65.78 &    66.08 &    69.03 &    22.75\\
\bottomrule\end{tabular}
}
\caption{Results for individual languages in the primary metric (CoNLL).
}
\label{tab:all-langs}
\end{table*}

\subsection{Evaluation of Zeros}

Table~\ref{tab:all-langs-zero} shows the performance of the systems on zero anaphora resolution on datasets with annotated zeros.
Let us start with a comparison of the \baseline{} and \baselinegz{} systems, which differ only in the nature of the empty nodes (predicted vs. gold).\footnote{
The gold empty nodes in the testset were not available to the participants,
 thus \baselinegz{} is not directly comparable with the other systems;
 it serves as a comparison with the previous year,
  when all empty nodes were gold.
}
It confirms that by moving to the realistic setup for zeros the task became much more challenging, illustrated by the performance drop in the F1 score by 5-19 points for most of the datasets.
\todo{performance of \baseline{} on \grcproiel{} is better. why?}
Note that for some datasets (\cspdt{}, \cspcedt{}, \plpcc{}) the task is so challenging that none of the systems was able to outperform \baselinegz{}.

If we ignore the results of \baselinegz{}, the winning  CorPipe-2stage system dominates the performance on zeros across most of the languages, being outperformed by the Ondfa systems on 4 datasets.
This correlates with the CoNLL scores across languages observed in Table~\ref{tab:all-langs}.
\todo{the only exception is \tritcc{}. Why?}
Interestingly, we observe huge disproportion in the performance changes between the winning system and the \baselinegz{} across the datasets of the same language.
Whereas the \baselinegz{} is better by 3 points on \cspdt{}, it is better by 14 points on \cspcedt{}.
Similarly, while the \baselinegz{} is worse by 2 points on \hukorkor{}, it is better by 19 points on \huszegedkoref{}.
It suggests significant differences in the guidelines for zero annotation across the datasets, even of the same language.

\begin{table*}\centering
\resizebox{\textwidth}{!}{

\begin{tabular}{@{}l r@{~~~}r@{~~~}r@{~~~}r@{~~~}r@{~~~}r@{~~~}r@{~~~}r@{~~~}r@{~~~}r@{}}\toprule
\bf system          & \rot\caancora & \rot\cspdt & \rot\cspcedt & \rot\cuproiel & \rot\esancora & \rot\grcproiel & \rot\hukorkor & \rot\huszegedkoref & \rot\plpcc & \rot\tritcc\\\midrule
CorPipe-2stage &       88  /      85  /      86   &       77  /      82  /      80   &       59  /      74  /      66   &  {\bf 75} / {\bf 78} / {\bf 76}  &  {\bf 90} / {\bf 92} / {\bf 91}  &  {\bf 84} / {\bf 88} / {\bf 86}  &       56  /      75  /      64   &  {\bf 83} /      68  /      75   &  {\bf 90} /      84  /      87   &  {\bf 83} /      80  /      82  \\
CorPipe       &       83  /      78  /      81   &       71  /      76  /      74   &       62  /      63  /      62   &  {\bf 75} /      74  /      75   &       84  /      84  /      84   &       79  /      83  /      81   &       55  /      74  /      63   &       71  /      68  /      70   &       85  /      78  /      82   &       70  /      68  /      69  \\
CorPipe-single &       81  /      77  /      79   &       72  /      72  /      72   &       63  /      58  /      60   &       75  /      72  /      73   &       83  /      83  /      83   &       80  /      77  /      78   &       52  /      71  /      60   &       72  /      65  /      68   &       83  /      75  /      79   &       66  /      60  /      63  \\
Ondfa         &  {\bf 88} / {\bf 86} / {\bf 87}  &       75  / {\bf 84} /      79   &       55  /      81  /      66   &       71  /      74  /      72   &       90  /      91  /      90   &       78  /      85  /      81   &       57  / {\bf 78} / {\bf 66}  &  {\bf 83} / {\bf 72} / {\bf 77}  &       90  /      83  /      86   &       82  / {\bf 82} / {\bf 82} \\
\baselinegz &       82  /      82  /      82   &  {\bf 82} /      84  / {\bf 83}  &  {\bf 78} / {\bf 82} / {\bf 80}  &       60  /      72  /      66   &       87  /      87  /      87   &       64  /      66  /      65   &  {\bf 60} /      65  /      62   &       53  /      59  /      56   &       89  / {\bf 86} / {\bf 87}  &       75  /      82  /      78  \\
\baseline     &       79  /      76  /      77   &       70  /      74  /      72   &       55  /      69  /      61   &       52  /      62  /      56   &       83  /      83  /      83   &       63  /      70  /      66   &       41  /      61  /      49   &       49  /      57  /      53   &       85  /      78  /      82   &       68  /      71  /      70  \\
DFKI-CorefGen &        0  /       0  /       0   &        0  /       0  /       0   &        0  /       0  /       0   &        0  /       0  /       0   &        0  /       0  /       0   &        0  /       0  /       0   &        0  /       0  /       0   &        0  /       0  /       0   &        0  /       0  /       0   &        0  /       0  /       0  \\
RitwikmishraFix &        0  /      50  /       0   &        0  /       0  /       0   &        0  /       0  /       0   &        0  /       0  /       0   &        0  /       0  /       0   &        0  /       0  /       0   &        0  /       0  /       0   &        0  /       0  /       0   &        0  /       0  /       0   &        0  /       0  /       0  \\
Ritwikmishra  &        0  /       0  /       0   &        0  /       0  /       0   &        0  /       0  /       0   &        0  /       0  /       0   &        0  /       0  /       0   &        0  /       0  /       0   &        0  /       0  /       0   &        0  /       0  /       0   &        0  /       0  /       0   &        0  /       0  /       0  \\
\midrule
\bestlast{}       &  93 / 92 / 92  &  91 / 92 / 92  &  87 / 88 / 87 & --  &  94 /      95  / 95  & -- &  82 /      89  / 85  &  88 /      70  /      78   &       75  /      69  /      72 & --  \\
\baselinelast{}     &       82  /      82  /      82   &       81  /      84  /      82   &       77  /      81  /      79   & -- &       87  /      88  /      87   &  -- &      60  /      68  /      64   &       61  /      57  /      59   &       50  / 80 /      62  & -- \\
\bottomrule\end{tabular}
}
\caption{Recall / Precision / F1 for anaphor-decomposable score of coreference resolution on zero anaphors across individual languages.
Only datasets containing anaphoric zeros are listed (\engum{} excluded as all zeros in its test set are non-anaphoric).
Note that these scores are directly comparable to neither the CoNLL score nor the supplementary scores calculated with respect to whole entities in Table~\ref{tab:secondary-metrics}.
}
\label{tab:all-langs-zero}
\end{table*}

Annual comparison of the results performed by baselines run in the gold zero setup (\baselinegz{} and \baselinelast{}) shows similar scores on zeros, which confirms that these baselines are comparable.
The only exception is \plpcc{}, on which \baselinegz{} improved by 25 percentage points.
This can be explained by the fixes in the CorefUD conversion pipeline from the source corpus that focused on zeros.
The annual comparison of relative improvements of the best systems over these baselines in terms of the zero anaphora score reveals that the improvements are much lower than they were last year, again confirming the more difficult nature of this year's setup for zeros.

\subsection{Further analysis}
Similarly to previous years, we provide several additional tables in the
appendices to shed more light on the differences between the submitted
systems.

Tables~\ref{tab:upos-entity}--\ref{tab:upos-mention} show results factorized
according to the different universal part of speech tags (UPOS) in the
mention heads. Table~\ref{tab:upos-entity} contains results on datasets where all
entities without any mention with a given UPOS as head were deleted.
Table~\ref{tab:upos-mention} contains results on datasets where all mentions
without a given UPOS as head were deleted, so these results may be a bit
misleading because e.g. the PRON column does not consider all pronominal
coreference, but only pronoun-to-pronoun coreference. An entity with one
pronoun and one noun mention is excluded from this table (because it becomes
a singleton after deleting noun or pronoun mentions and singletons are
excluded from the evaluation in these tables).

Tables \ref{tab:stats-entities}--\ref{tab:stats-details} show various
statistics on the entities and mentions in a concatenation of all the test
sets. Note that such statistics are mostly influenced by larger datasets.

Table \ref{tab:error-types} shows the distribution of error types
 based on the methodology of \citet{kummerfeld-klein-2013-error}
 and reveals that even systems with similar final F1 scores
 have different strengths and weaknesses.

\section{Conclusions and Future Work}
\label{sec:conclusions}

The paper summarizes the 2024 edition of the shared task on multilingual coreference resolution. Given that it is the third edition already, let us explore some generalizations.

First, the set of covered languages keeps growing: 11 languages in 2022, 13 languages in 2023, and 16 languages in 2024. 
Maintaining the pace of adding a few new languages each year seems realistic in the near future.

Second, in terms of the number of participating systems, the picture is mixed:
8 systems (5 teams) in 2022, 9 systems (7 teams) in 2023, and 6 systems (4 teams) in 2024. The relatively limited amount of participating teams can be partially attributed to the fact that the coreference resolution community is much smaller than e.g. the dependency parsing community. But still, it is an open question
why the shared task has not attracted more coreference research teams.

Third, although there is a great variance in performance both among
individual systems and across languages, the ordering of the systems remains
relatively stable. However, it is not straightforward to quantify the growth
of the state of the art along the individual shared task's editions;
comparing simply the absolute values of the primary score would not make
sense. The main reason is that the data collection gradually became bigger
and more diverse (e.g., by including typologically different languages, with
different scripts and different data sizes). At the same time, the task
itself differed slightly too, moving closer to real-world scenarios (by not
providing the participants with gold morphosyntactic annotation and gold zero
mentions in the input), which makes the task harder too.

One of the possible approaches to isolating the state-of-the-art growth trend
is to use the baseline system's performance as the point of reference because
the baseline's architecture remained unchanged throughout the three years.
The winner system outperformed the baseline's primary score by 21~\% relative
in 2022, by 31~\% relative in 2023, and by 39~\% relative in 2024. This
indicates that the task of multilingual coreference resolution is still in a
quickly progressing phase. We believe that the existence of this shared task
series was one of the most influential factors behind this growth.

For future iterations of this shared task, we plan to provide a sequence-to-sequence
(text-to-text) format for the training, evaluation and testing data. This new format
will be designed to simplify the use of large language models (LLMs) like GPT,
LLaMA, or Claude for the coreference resolution task.

The text-to-text format is particularly well suited for prompting approaches,
which have shown significant promise in various NLP tasks. By offering data in
this format, we aim to encourage more diverse approaches to the problem,
potentially leading to novel solutions and improved performance.

We will release this new data format alongside the existing CoNLL-U format,
giving participants the flexibility to choose the most suitable format for
their systems.

\section*{Acknowledgements}

This work has been supported
by Charles University Research Centre program No. 24/SSH/009,
Ministry of Education, Youth, and Sports of the Czech Republic, Project No.
LM2023062 LINDAT/CLARIAH-CZ,
Grant No. SGS-2022-016 Advanced methods of data processing and analysis,
and the Grant 20-16819X (LUSyD) of the Czech Science Foundation (GAČR).
We thank all the participants of the shared task for participating and for
providing brief descriptions of their systems. We also thank anonymous
reviewers for very useful remarks.

\newpage %

\bibliography{anthology,custom}
\bibliographystyle{acl_natbib}

\appendix
\clearpage
\onecolumn

\section{CorefUD 1.2 Details}
\label{sec:data-references}
\begin{tabular}{llll}
  Ancient Greek  & PROIEL       & \grcproiel{}      & \cite{Haug2008CreatingAP} \\
  Ancient Hebrew  & PTNK       & \hboptnk{}      & \cite{swanson-etal-2024-towards}\\
  Catalan    & AnCora       & \caancora{}      &  \cite{ancora,ancora-co} \\
  Czech      & PCEDT        & \cspcedt{}       & \cite{PCEDT2016coreference} \\
  Czech      & PDT          & \cspdt{}         & \cite{pdtconsolidated} \\
  English    & GUM          & \engum{}         & \cite{GUMdocumentation} \\
  English    & ParCorFull   & \enparcorfull{}  & \cite{ParCorFullScheme} \\
  English    & LitBank   & \enlitbank{}  & \cite{Bamman2019AnAD} \\
  French     & Democrat     & \frdemocrat{}    & \cite{democrat} \\
  German     & ParCorFull   & \deparcorfull{}  & \cite{ParCorFullScheme} \\
  German     & PotsdamCC    & \depotsdamcc{}   & \cite{bourgonje-stede-2020-potsdam} \\
  Hungarian  & KorKor       & \hukorkor{}      & \cite{korkor_coling} \\
  Hungarian  & SzegedKoref  & \huszegedkoref{} & \cite{szegedkoref2018} \\
  Lithuanian & LCC          & \ltlcc{}         & \cite{LithuanianScheme} \\
  Norwegian  & Bokmål NARC  & \nobokmaalnarc{} & \cite{maehlum2022narc} \\
  Norwegian  & Nynorsk NARC & \nonynorsknarc{} & \cite{maehlum2022narc} \\
  Old Church Slavonic     & PROIEL          & \cuproiel        & \cite{Haug2008CreatingAP} \\
  Polish     & PCC          & \plpcc{}         & \cite{PCC2013,bookOgrodniczuk} \\
  Russian    & RuCor        & \rurucor{}       & \cite{RuCorDialog} \\
  Spanish    & AnCora       & \esancora{}      & \cite{ancora,ancora-co} \\
  Turkish    & ITCC         & \tritcc          & \cite{pamay2018coref} \\
\end{tabular}

\section{CoNLL results by head UPOS}
\label{sec:stats-upos}

\begin{table}[H]\centering
\begin{tabular}{@{}l r@{~~~}r@{~~~}r@{~~~}r@{~~~}r@{~~~}r@{~~~}r@{~~~}r@{}}\toprule
\bf system         & \bf NOUN  & \bf PRON  & \bf PROPN & \bf DET   & \bf ADJ   & \bf VERB  & \bf ADV   & \bf NUM  \\\midrule
CorPipe-2stage     & \bf 70.23 & \bf 69.93 & \bf 76.23 &     49.20 &     42.45 &     33.64 &     28.70 & \bf 38.39 \\
CorPipe            &     69.06 &     69.66 &     75.07 &     52.35 & \bf 42.99 & \bf 35.02 & \bf 33.04 &     37.49 \\
CorPipe-single     &     66.69 &     66.90 &     71.72 & \bf 53.18 &     36.57 &     30.95 &     27.74 &     37.06 \\
Ondfa              &     66.79 &     66.54 &     69.18 &     49.08 &     33.61 &     26.90 &     29.98 &     34.18 \\
\baselinegz        &     48.49 &     55.58 &     52.18 &     32.39 &     25.05 &     11.34 &     17.67 &     28.09 \\
\baseline          &     46.77 &     49.73 &     51.51 &     33.08 &     23.65 &     10.83 &     16.89 &     26.66 \\
DFKI-CorefGen      &     30.49 &     33.97 &     31.54 &     18.50 &     10.11 &      2.72 &      8.56 &     10.57 \\
RitwikmishraFix    &     27.31 &     29.17 &     31.28 &     17.76 &     12.07 &      7.59 &      6.25 &      8.57 \\
Ritwikmishra       &     15.92 &     16.67 &     16.64 &     12.97 &      8.41 &      5.49 &      4.81 &      6.48 \\
\bottomrule\end{tabular}

\caption{CoNLL F1 score (head-match) evaluated only on entities with heads of a given
UPOS. In both the gold and prediction files we deleted some entities before
running the evaluation. We kept only entities with at least one mention with
a given head UPOS (universal part of speech tag). For the purpose of this
analysis,
 if the head node had deprel=flat children,
 their UPOS tags were considered as well,
 so for example in ``Mr./NOUN Brown/PROPN''
 both NOUN and PROPN were taken as head UPOS,
 so the entity with this mention will be reported in both columns NOUN and PROPN.
Otherwise, the CoNLL F1 scores are the same as in the primary metric,
 i.e. an unweighted average over all datasets, head-match, without singletons.
Note that when distinguishing entities into events and nominal entities,
 the VERB column can be considered as an approximation of the performance on events.
One of the limitations of this approach is that copula is not treated as head in the Universal Dependencies,
 so, e.g., phrase \textit{She is nice} is not considered for the VERB column,
 but for the ADJ column (head of the phrase is \textit{nice}).
}
\label{tab:upos-entity}
\end{table}

\begin{table}[H]\centering
\begin{tabular}{@{}l r@{~~~}r@{~~~}r@{~~~}r@{~~~}r@{~~~}r@{~~~}r@{~~~}r@{}}\toprule
\bf system         & \bf NOUN  & \bf PRON  & \bf PROPN & \bf DET   & \bf ADJ   & \bf VERB  & \bf ADV   & \bf NUM  \\\midrule
CorPipe-2stage     & \bf 60.43 & \bf 60.00 & \bf 61.33 & \bf 49.58 & \bf 47.09 & \bf 47.07 & \bf 48.05 & \bf 46.82 \\
CorPipe            &     59.37 &     58.26 &     60.22 &     47.00 &     44.31 &     43.99 &     44.53 &     44.31 \\
CorPipe-single     &     55.50 &     55.25 &     54.64 &     43.08 &     40.28 &     39.77 &     39.77 &     39.91 \\
Ondfa              &     57.22 &     54.58 &     56.04 &     44.21 &     41.65 &     41.28 &     41.34 &     41.42 \\
\baselinegz        &     38.50 &     45.45 &     39.85 &     28.88 &     26.23 &     26.06 &     26.29 &     26.06 \\
\baseline          &     37.30 &     39.46 &     39.46 &     27.84 &     25.52 &     25.12 &     25.56 &     25.30 \\
DFKI-CorefGen      &     20.99 &     26.05 &     22.71 &     16.68 &     14.24 &     14.04 &     14.46 &     14.20 \\
RitwikmishraFix    &     25.26 &     26.08 &     25.53 &     18.06 &     17.01 &     16.27 &     16.43 &     16.49 \\
Ritwikmishra       &     14.29 &     14.05 &     12.74 &     10.38 &      9.56 &      8.89 &      9.12 &      9.13 \\
\bottomrule\end{tabular}
\caption{CoNLL F1 score (head-match) evaluated only on mentions with heads of a given UPOS.
In both the gold and prediction files we deleted some mentions before running the evaluation.
We kept only mentions with a given head UPOS (again considering also deprel=flat children).}
\label{tab:upos-mention}
\end{table}

\section{Statistics of the submitted systems on concatenation of all test sets}
\label{sec:stats-concat}
The systems are sorted alphabetically in tables in this section.
The predictions of the Ritwikmishra system were not valid CoNLL-U
 and thus are excluded in these tables
 (the script collecting the statistics failed),
 see the numbers of the RitwikmishraFix system instead.

\begin{table}[H]\centering
\begin{tabular}{@{}l rrrr rrrrr@{}}\toprule
                    & \MC{4}{entities}                  & \MC{5}{distribution of lengths}      \\\cmidrule(lr){2-5}\cmidrule(l){6-10}
system              &   total & per 1k & \MC{2}{length} &     1 &     2 &     3 &     4 &   5+ \\\cmidrule(lr){4-5}
                    &   count &  words &    max &  avg. &  [\%] &  [\%] &  [\%] &  [\%] & [\%] \\\midrule
gold                &  47,680 &    102 &    509 &   2.2 &  61.0 &  21.9 &   6.8 &   3.3 &   7.0 \\
\baseline           &  15,168 &     33 &    154 &   3.9 &   0.0 &  57.4 &  17.0 &   7.7 &  17.9 \\
\baselinegz         &  15,534 &     33 &    154 &   3.9 &   0.0 &  57.4 &  17.1 &   7.8 &  17.7 \\
CorPipe             &  49,943 &    107 &    288 &   2.1 &  62.1 &  20.5 &   7.1 &   3.3 &   7.0 \\
CorPipe-2stage      &  49,980 &    107 &    299 &   2.1 &  62.4 &  20.7 &   6.9 &   3.2 &   6.8 \\
CorPipe-single      &  50,179 &    108 &    573 &   2.1 &  62.4 &  20.2 &   7.0 &   3.4 &   7.1 \\
DFKI-CorefGen       &  33,188 &     71 &    191 &   2.1 &  70.3 &  14.9 &   5.7 &   2.6 &   6.4 \\
Ondfa               &  48,739 &    105 &    203 &   2.1 &  63.5 &  20.1 &   6.4 &   3.1 &   6.9 \\
RitwikmishraFix     &   6,703 &     14 &    637 &   3.5 &  29.2 &  37.3 &  13.0 &   6.0 &  14.5 \\
\bottomrule\end{tabular}
\caption{Statistics on coreference entities.
The total number of entities and the average number of entities per 1000 tokens in the running text.
The maximum and average entity ``length'',
 i.e., the number of mentions in the entity.
Distribution of entity lengths (singletons have length = 1).
The four best systems (CorPipe* and Ondfa) have the statistics similar to the gold data
 (although they all slightly overgenerate,
  i.e. predicts more entities than in the gold data).
The remaining systems undergenerate and the two baselines and RitwikmishraFix also
 predict on average longer entities (i.e. with more mentions) than in the gold data.
}
\label{tab:stats-entities}
\end{table}

\begin{table}[H]\centering
\begin{tabular}{@{}l rrrr rrrrrr@{}}\toprule
                    & \MC{4}{mentions}                  & \MC{6}{distribution of lengths}              \\\cmidrule(lr){2-5}\cmidrule(l){6-11}
system              &   total & per 1k & \MC{2}{length} &     0 &     1 &     2 &     3 &     4 &   5+ \\\cmidrule(lr){4-5}
                    &   count &  words &    max &  avg. &  [\%] &  [\%] &  [\%] &  [\%] &  [\%] & [\%] \\\midrule
gold                &  74,305 &    159 &    100 &   2.9 &  12.6 &  44.0 &  18.1 &   7.3 &   3.6 &  14.3 \\
\baseline           &  59,859 &    128 &     27 &   2.1 &  14.8 &  47.4 &  17.8 &   6.6 &   3.1 &  10.2 \\
\baselinegz         &  61,277 &    131 &     27 &   2.1 &  14.8 &  47.0 &  17.9 &   6.8 &   3.1 &  10.5 \\
CorPipe             &  74,076 &    159 &    100 &   2.9 &  12.5 &  44.6 &  18.1 &   7.3 &   3.5 &  14.0 \\
CorPipe-2stage      &  73,239 &    157 &    116 &   2.8 &  12.4 &  44.9 &  18.1 &   7.3 &   3.5 &  13.7 \\
CorPipe-single      &  75,350 &    162 &    145 &   2.8 &  12.9 &  44.3 &  18.1 &   7.4 &   3.5 &  13.8 \\
DFKI-CorefGen       &  44,731 &     96 &     65 &   2.6 &   0.0 &  57.4 &  20.5 &   7.0 &   3.3 &  11.8 \\
Ondfa               &  71,531 &    153 &     22 &   1.1 &  12.3 &  82.1 &   2.1 &   1.1 &   0.5 &   2.0 \\
RitwikmishraFix     &  21,458 &     46 &     16 &   1.5 &   0.0 &  66.5 &  22.6 &   7.0 &   2.1 &   1.8 \\
\bottomrule\end{tabular}

\caption{Statistics on non-singleton mentions.
The total number of mentions and the average number of
mentions per 1000 words of running text. The maximum and average mention length, i.e., the number of nonempty nodes (words) in the mention. Distribution of mention lengths (zeros have length = 0).
The four best systems (CorPipe* and Ondfa) generate a similar number of non-singleton mentions as in the gold data
(although last year, the three best systems overgenerated mentions).
The average length of mentions predicted by Ondfa is notably lower than in the gold data
 because Ondfa predicted single-word mentions only in all datasets except for \cspcedt{} and \ltlcc{}.
No system predicts long mentions (4 and 5+ words) more frequently than in the gold data,
 although CorPipe is near to the gold distribution.
}
\label{tab:stats-mentions-nonsingleton}
\end{table}

\begin{table}[H]\centering
\begin{tabular}{@{}l rrrr rrrrrr@{}}\toprule
                    & \MC{4}{mentions}                  & \MC{6}{distribution of lengths}              \\\cmidrule(lr){2-5}\cmidrule(l){6-11}
system              &   total & per 1k & \MC{2}{length} &     0 &     1 &     2 &     3 &     4 &   5+ \\\cmidrule(lr){4-5}
                    &   count &  words &    max &  avg. &  [\%] &  [\%] &  [\%] &  [\%] &  [\%] & [\%] \\\midrule
gold                &  29,087 &     62 &     81 &   3.4 &   1.8 &  30.8 &  24.7 &  13.7 &   7.5 &  21.6 \\
\baseline           &       0 &      0 &      0 &   0.0 &   0.0 &   0.0 &   0.0 &   0.0 &   0.0 &   0.0 \\
\baselinegz         &       0 &      0 &      0 &   0.0 &   0.0 &   0.0 &   0.0 &   0.0 &   0.0 &   0.0 \\
CorPipe             &  31,030 &     67 &    163 &   3.5 &   2.0 &  29.7 &  25.7 &  13.8 &   7.6 &  21.4 \\
CorPipe-2stage      &  31,164 &     67 &    163 &   3.5 &   2.1 &  29.9 &  25.9 &  13.9 &   7.5 &  20.7 \\
CorPipe-single      &  31,309 &     67 &     93 &   3.5 &   1.7 &  29.8 &  25.6 &  13.9 &   7.6 &  21.4 \\
DFKI-CorefGen       &  23,342 &     50 &     71 &   2.9 &   0.0 &  35.5 &  28.5 &  13.4 &   6.7 &  15.9 \\
Ondfa               &  30,971 &     66 &     19 &   1.0 &   2.1 &  96.3 &   0.5 &   0.3 &   0.2 &   0.5 \\
RitwikmishraFix     &   1,959 &      4 &     13 &   1.8 &   0.0 &  45.6 &  40.0 &  10.4 &   2.6 &   1.4 \\
\bottomrule\end{tabular}
\caption{Statistics on singleton mentions.
See the caption of Table~\ref{tab:stats-mentions-nonsingleton} for details.
The two baseline systems do not attempt to predict singletons at all.
Interestingly, last year all systems predicted 7--9 times less singletons than in the gold data.
This year, the four best systems (CorPipe* and Ondfa) predict slightly more singletons than in the gold data.
Note that singletons are not annotated in all the (gold) datasets.
}
\label{tab:stats-mentions-singleton}
\end{table}

\begin{table}[H]\centering
\begin{tabular}{@{}l @{}r@{~}r@{~}r @{~}r@{~}r@{~}r@{~}r@{~}r@{~}r@{~}r@{~}r@{~}r@{~}r@{}}\toprule
                    & \MC{3}{mention type [\%]}    & \MC{10}{distribution of head UPOS [\%]}      \\\cmidrule(lr){2-4}\cmidrule(l){5-14}
system              & w/empty & w/gap & non-tree
                                            &  NOUN &  PRON & PROPN &   DET &   ADJ &  VERB &   ADV &   NUM &   \_~ & other \\\midrule
gold                &  14.7 &   0.7 &   1.6 &  40.2 &  28.6 &  14.7 &   6.7 &   2.5 &   2.2 &   1.1 &   0.5 &   2.8 &   0.6 \\    
\baseline           &  15.9 &   0.0 &   1.6 &  36.6 &  20.3 &  15.6 &   7.5 &   2.3 &   0.9 &   1.1 &   0.3 &  14.9 &   0.5 \\    
\baselinegz         &  16.0 &   0.0 &   1.7 &  37.1 &  31.4 &  15.4 &   7.5 &   2.2 &   1.0 &   1.1 &   0.4 &   3.4 &   0.5 \\    
CorPipe             &  14.0 &   0.0 &   1.8 &  40.4 &  19.0 &  14.9 &   6.9 &   2.3 &   1.8 &   1.1 &   0.4 &  12.5 &   0.7 \\    
CorPipe-2stage      &  13.8 &   0.0 &   1.9 &  40.3 &  19.1 &  15.0 &   6.9 &   2.4 &   1.6 &   1.1 &   0.5 &  12.5 &   0.6 \\    
CorPipe-single      &  14.4 &   0.0 &   1.8 &  40.5 &  18.8 &  14.7 &   6.8 &   2.3 &   1.7 &   1.1 &   0.5 &  12.9 &   0.6 \\    
DFKI-CorefGen       &   0.0 &   0.0 &   3.9 &  40.7 &  27.8 &  16.3 &  10.0 &   1.4 &   1.0 &   1.2 &   0.4 &   0.0 &   1.2 \\    
Ondfa               &  12.6 &   0.0 &   0.2 &  40.6 &  19.2 &  14.8 &   6.9 &   2.5 &   1.6 &   1.2 &   0.5 &  12.3 &   0.5 \\    
RitwikmishraFix     &   0.1 &   0.0 &   0.8 &  28.9 &  31.3 &  27.7 &   5.7 &   1.8 &   2.3 &   0.8 &   0.8 &   0.0 &   0.6 \\    
\bottomrule\end{tabular}
\caption{Detailed statistics on non-singleton mentions.
The left part of the table shows the percentage of:
 mentions with at least one empty node (w/empty);
 mentions with at least one gap, i.e. discontinuous mentions (w/gap);
 and non-treelet mentions, i.e. mentions not forming a connected subgraph (catena) in the dependency tree (non-tree).
Note that these three types of mentions may be overlapping.
We can see that none of the systems attempts to predict discontinuous mentions.
DFKI-CorefGen has a notably higher percentage (3.9\%) of non-treelet mention spans.
The right part of the table shows the distribution of mentions
  based on the universal part-of-speech tag (UPOS) of the head word.
Note that this distribution has to be interpreted with the total number of non-singleton mentions predicted (as reported in Table~\ref{tab:stats-mentions-nonsingleton}) in mind.
For example, 31.4\% of non-singleton mentions predicted by \baselinegz{} are pronominal (head=PRON),
  while there are only 28.6\% of pronominal non-singleton mentions in the gold data.
However, \baselinegz{} predicts actually less pronominal non-singleton mentions (61277*31.4\%=19241) than in the gold data (74305*28.6\%=21251).
Note that the same word may be assigned a different UPOS tag in the predicted and gold data
(in case of empty nodes or if the gold data includes manual annotation).
The empty UPOS tag (\_) is present only in the empty nodes
and none of the systems attempts to predict the actual UPOS tag of empty nodes
(they all keep the empty tag from the baseline predictor of empty nodes,
although about 78\% of the empty nodes in the gold devset are pronouns).
}
\label{tab:stats-details}
\end{table}

\begin{table}[H]
\resizebox{\textwidth}{!}{%
\newcommand{\partialbar}[2]{%
    \begin{tikzpicture}
        \draw[fill=black] (0,0) rectangle (#1,0.3);
        \draw[fill=white] (#1,0) rectangle (#2,0.3);
    \end{tikzpicture}%
}

\begin{tabular}{l m{35pt} m{35pt} m{45pt} m{50pt} m{40pt} m{50pt} m{40pt}}
\toprule
System & Span Errors & Extra Entity Errors & Extra Mention Errors & Conflated Entities Errors & Missing Entity Errors & Missing Mention Errors & Divided Entity Errors \\ \midrule
\baseline  & \partialbar{0.07}{1} & \partialbar{0.65}{1} & \partialbar{0.41}{1} & \partialbar{0.7}{1} & \partialbar{0.36}{1} & \partialbar{0.52}{1} & \partialbar{0.91}{1}\\
\baselinegz  & \partialbar{0.06}{1} & \partialbar{0.72}{1} & \partialbar{0.33}{1} & \partialbar{0.75}{1} & \partialbar{0.36}{1} & \partialbar{0.43}{1} & \partialbar{0.99}{1}\\
CorPipe  & \partialbar{0.13}{1} & \partialbar{0.95}{1} & \partialbar{0.92}{1} & \partialbar{0.52}{1} & \partialbar{0.17}{1} & \partialbar{0.51}{1} & \partialbar{0.69}{1}\\
CorPipe-2stage  & \partialbar{0.1}{1} & \partialbar{0.83}{1} & \partialbar{0.43}{1} & \partialbar{0.48}{1} & \partialbar{0.16}{1} & \partialbar{0.27}{1} & \partialbar{0.67}{1}\\
CorPipe-single  & \partialbar{0.14}{1} & \partialbar{1.0}{1} & \partialbar{1.0}{1} & \partialbar{0.55}{1} & \partialbar{0.17}{1} & \partialbar{0.51}{1} & \partialbar{0.72}{1}\\
DFKI-CorefGen  & \partialbar{0.11}{1} & \partialbar{0.62}{1} & \partialbar{0.62}{1} & \partialbar{1.0}{1} & \partialbar{0.55}{1} & \partialbar{1.0}{1} & \partialbar{1.0}{1}\\
Ondfa  & \partialbar{1.0}{1} & \partialbar{0.85}{1} & \partialbar{0.43}{1} & \partialbar{0.52}{1} & \partialbar{0.2}{1} & \partialbar{0.3}{1} & \partialbar{0.64}{1}\\
RitwikmishraFix  & \partialbar{0.07}{1} & \partialbar{0.48}{1} & \partialbar{0.26}{1} & \partialbar{0.18}{1} & \partialbar{1.0}{1} & \partialbar{0.39}{1} & \partialbar{0.21}{1}\\
\midrule
\textit{Most Errors} & 22120 & 2711 & 10709 & 3570 & 15095 & 20088 & 2493 \\ \bottomrule
\end{tabular}
}
\caption{Distribution of error types based on the methodology of \citet{kummerfeld-klein-2013-error}. By gradually transforming the prediction files into gold data, we can classify several types of transformations, which then map to types of errors. The number in the last row is the maximal total number of errors (summed over all datasets) of the given type, that any of the predictions made. The partially filled bars display the percentage of the maximal number of errors in the given column. The table should be viewed column-wise to compare individual prediction systems.
The Span Errors column shows once again that Ondfa does not attempt to predict the whole span (only the head).
CorPipe-single and CorPipe are the two worst systems in the number of Extra Entity and Extra Mention errors.
However, according to Table~\ref{tab:secondary-metrics}, these systems have recall as high as precision,
while other systems (e.g. Ondfa) have recall much lower;
thus the high number of extra entities and mentions seems to be a good trade-off.
Interestingly, CorPipe-2stage has the same recall as CorPipe (in almost all metric),
but a slightly higher precision in Table~\ref{tab:secondary-metrics},
which corresponds to the relatively lower number of Extra Entity and especially Extra Mention errors.
}
\label{tab:error-types}
\end{table}

\end{document}